\documentclass[conference]{IEEEtran}
\IEEEoverridecommandlockouts
\usepackage{amsmath,amssymb,amsfonts}
\usepackage{algorithmic}
\usepackage{graphicx}
\usepackage{textcomp}
\usepackage{xcolor}
\usepackage{changepage} % for adjustwidth environment
\usepackage{pgfplots}
\usepackage{subcaption}
\pgfplotsset{compat=1.17}
\usepackage{verbatim}
\def\BibTeX{{\rm B\kern-.05em{\sc i\kern-.025em b}\kern-.08em
    T\kern-.1667em\lower.7ex\hbox{E}\kern-.125emX}}
\usepackage[numbers,sort&compress]{natbib}
\usepackage{hyperref}
\hypersetup{
colorlinks   = true,
citecolor    = blue,
urlcolor    =  blue
}
\usepackage{url}
\usepackage{multirow}
\usepackage{hhline}
\usepackage{makecell}
\usepackage{tabularx}
\newcolumntype{Y}{>{\centering\arraybackslash}X}
\newcolumntype{Z}{>{\raggedleft\arraybackslash}X}

\newcommand{\footURL}[1]{\footnote{\url{#1}}}

\makeatletter
\newcommand\footnoteref[1]{\protected@xdef\@thefnmark{\ref{#1}}\@footnotemark}
\makeatother

\begin{document}

\title{Aspect-Based Sentiment Analysis Techniques: A Comparative Study%\\
%{\footnotesize \textsuperscript{*}Note: Sub-titles are not captured in Xplore and should not be used}
}

\author{
\IEEEauthorblockN{Dineth Jayakody\IEEEauthorrefmark{1}\IEEEauthorrefmark{2}, Koshila Isuranda\IEEEauthorrefmark{1}\IEEEauthorrefmark{2}, A V A Malkith\IEEEauthorrefmark{1}\IEEEauthorrefmark{2}, Nisansa de Silva\IEEEauthorrefmark{3},\\ Sachintha Rajith Ponnamperuma\IEEEauthorrefmark{4}, G G N Sandamali\IEEEauthorrefmark{1}\IEEEauthorrefmark{5}, K L K Sudheera\IEEEauthorrefmark{1}\IEEEauthorrefmark{5}}
\IEEEauthorblockA{\IEEEauthorrefmark{1}\textit{Department of Electrical and Information Engineering}, \textit{University of Ruhuna}\\
\IEEEauthorrefmark{2}\texttt{\{jayakody\_ds\_e21,isuranda\_mak\_e21, malkith\_ava\_e21\}@engug.ruh.ac.lk}}
\IEEEauthorrefmark{5}\texttt{\{nadeesha, kushan\}@eie.ruh.ac.lk}
\IEEEauthorblockA{\IEEEauthorrefmark{3}\textit{Department of Computer Science \& Engineering}, \textit{University of Moratuwa}\\
\IEEEauthorrefmark{3}\texttt{NisansaDdS@cse.mrt.ac.lk}}
\IEEEauthorblockA{\IEEEauthorrefmark{4}\textit{Emojot Inc.}\\
\IEEEauthorrefmark{4}\texttt{sachintha@emojot.com}}
}

%\author{\IEEEauthorblockN{Anonymous Authors}\IEEEauthorblockA{\textit{Anonymous Affiliations} \\ \textit{Anonymous Affiliations}\\ \texttt{Anonymous emails}}}

\maketitle

\begin{abstract}
Since the dawn of the digitalisation era, customer feedback and online reviews are unequivocally major sources of insights for businesses. Consequently, conducting comparative analyses of such sources has become the de facto modus operandi of any business that wishes to give itself a competitive edge over its peers and improve customer loyalty. Sentiment analysis is one such method instrumental in gauging public interest, exposing market trends, and analysing competitors. While traditional sentiment analysis focuses on overall sentiment, as the needs advance with time, it has become important to explore public opinions and sentiments on various specific subjects, products and services mentioned in the reviews on a finer-granular level. To this end, Aspect-based Sentiment Analysis (ABSA), supported by advances in Artificial Intelligence (AI) techniques which have contributed to a paradigm shift from simple word-level analysis to tone and context-aware analyses, focuses on identifying specific aspects within the text and determining the sentiment associated with each aspect. %As such, a clearer understanding of the opinions with detailed insights is provided by ABSA. 
%Due to the complexity of morphologically rich languages, complex algorithms are needed to extract and analyze specific aspects within the text. Therefore, performing ABSA accurately still remains a challenge.
In this study, we compare several deep-NN methods for ABSA on two benchmark datasets (Restaurant-14 and Laptop-14) and found that \textit{FAST LSA} obtains the best overall results of 87.6\% and 82.6\% accuracy but does not pass \textit{LSA+DeBERTa} which reports 90.33\% and 86.21\% accuracy respectively.
\end{abstract}


\section{Introduction}
Social media and other online platforms have enjoyed exponential growth, which in turn has created an unprecedented abundance of user-generated content. However, conversely, this has added a de facto expectation on businesses to understand the user sentiments expressed in these texts if they intend to make informed decisions and enhance customer satisfaction.
% With the exponential growth of online platforms and social media, the volume of user-generated content has reached unprecedented levels. Understanding user sentiments expressed in these texts is crucial for businesses to make informed decisions and enhance customer satisfaction. 
Aspect-based sentiment analysis (ABSA) has emerged as a valuable technique in Natural Language Processing (NLP) to analyze opinions at a finer granular level by identifying sentiment towards specific aspects or features within a given domain~\cite{mudalige2020sigmalaw,rajapaksha2021sigmalaw,jayasinghe2021party,rajapaksha2020rule,rajapaksha2022sigmalaw,samarawickrama2022legal}.
%Aspect-based sentiment analysis (ABSA) has emerged as a valuable application of Natural Language Processing (NLP), allowing for a more detailed analysis of opinions by identifying sentiment towards specific aspects or features within a given domain.
In this paper, we focus on domain-specific ABSA~\cite{rajapaksha2020rule,rajapaksha2021sigmalaw,jayasinghe2021party}, particularly focusing on its application in analyzing customer reviews, which provides valuable feedback for businesses to improve their products or services.
% In this paper, we present our research on domain-specific aspect-based sentiment analysis, focusing on its application in analyzing customer reviews. 
%Our collaboration with Emojot (Pvt) Ltd. has provided us with access to their customer review platform, enabling us to deploy and evaluate ABSA techniques in real-world scenarios.

In our initial analysis, we present the accuracy levels of various ABSA models using the benchmark 2014 SemEval restaurant and laptop datasets~\cite{pontiki-etal-2014-semeval}. The results are tabulated to provide a comparative view of their performance in sentiment analysis tasks. Following this assessment, we proceed to explore avenues for improving model performance through fine-tuning and testing on the same dataset. Through the fine-tuning process, our objective is to adapt these pre-trained models to the specific variations of the domain under consideration, thereby enhancing their effectiveness in capturing sentiment expressions related to various aspects contained within customer reviews. Specifically, we focus on the \texttt{Llama 2} model, which utilizes parameter-efficient techniques enabled by the \texttt{QLora} architecture, as well as the \texttt{FAST\_LSA\_T\_V2} model, integrated into the \texttt{PyABSA} framework. Additionally, we investigate transformer pre-trained models within the context of the SetFit framework. Leveraging this framework, we experiment with hybrid models by connecting different transformer models and testing them on the aforementioned dataset. 

Further elaboration on these models and their fine-tuning methodologies will be provided in subsequent sections of this paper, where we look into their architectures, techniques, and experimental results in greater detail.

\section{Literature Review}
%Our literature review encompasses an exploration of various models that have demonstrated notable performance, achieving mean accuracies exceeding 80\% on the SemEval 2014 restaurant and laptop datasets.

\citet{rietzler2019adapt} undertook a comprehensive study focusing on Aspect-Target Sentiment Classification (ATSC) within ABSA, presenting a novel two-step approach. Their methodology involved domain-specific fine-tuning of \texttt{BERT}~\cite{devlin2018bert} language models followed by task-specific fine-tuning, resulting in an accuracy of approximately 84.06\% with the \texttt{BERT-ADA} model which surpassed the performance of baseline models such as vanilla \texttt{BERT-base} and \texttt{XLNet-base}~\cite{yang2019xlnet}. The success of this model underscores the significance of domain-specific considerations for improving model robustness and performance in real-world applications.

%This study not only underscores the effectiveness of advanced language models in sentiment analysis tasks but also highlights the significance of domain-specific considerations for improving model robustness and performance in real-world applications.

Subsequent studies by \citet{karimi2021adversarial} and \citet{bai2020investigating} explored alternative approaches utilizing BERT-based models for ABSA. \citet{karimi2021adversarial} introduced adversarial training to enhance ABSA performance, leveraging artificial data generation through adversarial processes. Their BERT Adversarial Training \texttt{(BAT)} architecture surpassed both general-purpose \texttt{BERT} and domain-specific post-trained \texttt{BERT} (\texttt{BERT-PT}) models in ABSA tasks, without the need for extensive manual labelling. Similarly, \citet{karimi2020improving} introduced two novel modules, Parallel Aggregation and Hierarchical Aggregation, to augment ABSA using \texttt{BERT}. These modules aimed to enhance Aspect Extraction (AE) and Aspect Sentiment Classification (ASC) tasks, yielding superior performance compared to post-trained vanilla \texttt{BERT}.
%Following groundbreaking work of \citet{rietzler2019adapt}, subsequent studies by \citet{karimi2021adversarial} and \citet{bai2020investigating} explored alternative approaches utilizing BERT-based models for Aspect-Based Sentiment Analysis (ABSA). \citet{karimi2021adversarial} introduced adversarial training as an innovative method to enhance ABSA performance, leveraging artificial data generation through adversarial processes. Their \texttt{(BERT)} Adversarial Training \texttt{(BAT)} architecture surpassed both general-purpose \texttt{BERT} and domain-specific post-trained \texttt{BERT} (\texttt{BERT-PT}) models in ABSA tasks, showcasing its efficacy in sentiment analysis without the need for extensive manual labelling. Similarly, \citet{karimi2020improving} introduced two novel modules, Parallel Aggregation and Hierarchical Aggregation, to augment ABSA using \texttt{BERT}. These modules aimed to enhance Aspect Extraction (AE) and Aspect Sentiment Classification (ASC) tasks, yielding superior performance compared to post-trained vanilla \texttt{BERT}.

Introducing a multi-task learning model for ABSA, \citet{yang2021multi} achieved an accuracy of 86.60\% with their \texttt{LCF-ATEPC} model. Meanwhile, \citet{dai2021does} explored the potential of pre-trained models (\texttt{PTMs}), particularly the \texttt{RoBERTa}~\cite{cer2017semeval} model in ABSA tasks but could not surpass the performance of \texttt{LCF-ATEPC} model. The superior performance of the \texttt{LCF-ATEPC} highlights the efficacy of multi-task learning approaches in ABSA, showcasing the importance of considering aspect term extraction (ATE) alongside polarity classification for comprehensive sentiment analysis.
%Introducing a pioneering multi-task learning model for Aspect-Based Sentiment Analysis (ABSA) tasks, \citet{yang2021multi} achieved an impressive accuracy of more than 86.60\% with their \texttt{LCF-ATEPC} model. Meanwhile, \citet{dai2021does} explored the potential of pre-trained models (\texttt{PTMs}), particularly the \texttt{RoBERTa} model, in ABSA tasks but could not surpass the performance of \texttt{LCF-ATEPC} model of \citet{yang2021multi}. The study of \citet{dai2021does} revealed insights into the effectiveness of fine-tuning \texttt{PTMs} for ABSA tasks, suggesting promising directions for future research in sentiment analysis techniques. However, the superior performance of \texttt{LCF-ATEPC} highlights the efficacy of multi-task learning approaches in ABSA, showcasing the importance of considering aspect term extraction alongside polarity classification for comprehensive sentiment analysis.

\texttt{DeBERTa}~\cite{he2020deberta} (Decoding-enhanced \texttt{BERT} with Disentangled Attention)-based models were explored by \citet{silva2021aspect} and \citet{yang2021improving}, introducing \texttt{ABSA-DeBERTa} and \texttt{LSA+DeBERTa-V3-Large}, respectively. \citet{silva2021aspect} delved into disentangled learning to enhance \texttt{BERT}-based representations in ABSA. They separated syntactic and semantic features, showcasing the improvement in ABSA task performance through the incorporation of disentangled attention. This enabled the isolation of position and content vectors, potentially enhancing model performance by focusing on syntactic and semantic aspects separately. On the other hand, \citet{yang2021improving} introduced a novel perspective in aspect-based sentiment classification (ABSC) by emphasizing the significance of aspect sentiment coherency.
%  \citet{silva2021aspect} and \citet{yang2021improving} explored the utilization of \texttt{DeBERTa}-based~\cite{he2020deberta} models, introducing \texttt{ABSA-DeBERTa} and \texttt{LSA+DeBERTa-V3-Large}, respectively. \citet{silva2021aspect} delved into disentangled learning to enhance \texttt{BERT-based} representations in Aspect-Based Sentiment Analysis (ABSA). Leveraging the \texttt{DeBERTa} (Decoding-enhanced \texttt{BERT} with Disentangled Attention) model, they separated syntactic and semantic features, showcasing the improvement in ABSA task performance through the incorporation of disentangled attention. This approach enables the isolation of position and content vectors, potentially enhancing model performance by focusing on syntactic and semantic aspects separately. On the other hand, \citet{yang2021improving} introduced a novel perspective in aspect-based sentiment classification (ABSC) by emphasizing the significance of aspect sentiment coherency.
Subsequently, \citet{xing2022understand} and \citet{zhang2022towards} also explored the utilization of \texttt{BERT}-based models, introducing \texttt{KaGRMN-DSG} (Knowledge-aware Gated Recurrent Memory Network with Dual Syntax Graph Modeling) and \texttt{DPL-BERT}, respectively.
%. \citet{xing2022understand} addressed challenges in ABSC with their \texttt{KaGRMN-DSG}  model. 
However, despite these advancements, neither \texttt{KaGRMN-DSG} nor \texttt{DPL-BERT} could surpass the accuracy achieved by the \texttt{LCF-ATEPC}~\cite{yang2021multi} model.
%In 2022, \citet{xing2022understand} and \citet{zhang2022towards} also explored the utilization of \texttt{BERT-based} models, introducing \texttt{KaGRMN-DSG} and \texttt{DPL-BERT}, respectively. \citet{xing2022understand} addressed challenges in Aspect-level Sentiment Classification (ASC) with their \texttt{KaGRMN-DSG} (Knowledge-aware Gated Recurrent Memory Network with Dual Syntax Graph Modeling) model. However, despite these advancements, neither \texttt{KaGRMN-DSG} nor \texttt{DPL-BERT} could surpass the accuracy achieved by the \texttt{LCF-ATEPC} model introduced by \citet{yang2021multi}.

In \autoref{tab:litAccura} we summarize the accuracies of the models the relevant studies in the literature have reported for the benchmark SemEval~\cite{pontiki-etal-2014-semeval} Restaurant (\textit{Res-14}) and Laptop (\textit{Lap-14}) datasets. The accuracies range from 82.69\% to 88.27\%, showcasing the varying degrees of success in sentiment analysis across different approaches.
%Comparing the accuracies of the models listed in the \autoref{tab1} reveals interesting insights into their performance on the SemEval 2014 Restaurant and Laptop datasets. The accuracies range from 82.69\% to 88.27\%, showcasing the varying degrees of success in sentiment analysis across different approaches.

\begin{table}[htbp]
\caption{Accuracies of models on the SemEval 2014~\cite{pontiki-etal-2014-semeval} benchmark}
\label{tab:litAccura}
\begin{center}
%\begin{tabular}{|l|c|c|c|}
\begin{tabularx}{0.45\textwidth}{lYYY}
\hline
\multirow{2}{*}{\textbf{Model}} & \multicolumn{3}{c}{\textbf{Accuracy}} \\
\hhline{~---}
 & \textbf{Res-14} & \textbf{Lap-14} & \textbf{Mean}  \\ \hline
BAT~\cite{karimi2021adversarial} & 86.03 & 79.35 & 82.69 \\ %\hline
PH-SUM~\cite{karimi2020improving} & 86.37 & 79.55 & 82.96 \\ %\hline
RGAT+~\cite{bai2020investigating}  & 86.68 & 80.94 & 83.81 \\ %\hline
BERT-ADA~\cite{rietzler2019adapt} & 87.89 & 80.23 & 84.06  \\ %\hline
KaGRMN-DSG~\cite{xing2022understand}  & 87.35 & 81.87 & 84.61 \\ %\hline
RoBERTa+MLP~\cite{dai2021does}  & 87.37 & 83.78 & 85.58 \\ %\hline
DPL-BERT~\cite{zhang2022towards}  & 89.54 & 81.96 & 85.75 \\ %\hline
ABSA-DeBERTa~\cite{silva2021aspect}  & 89.46 & 82.76 & 86.11 \\ %\hline
LCF-ATEPC~\cite{yang2021multi} & 90.18 & 83.02 & 86.60  \\ %\hline
LSA+DeBERTa-V3-Large~\cite{yang2021improving}  & \textbf{90.33} & \textbf{86.21} & \textbf{88.27} \\ \hline
%\end{tabular}
\end{tabularx}
\end{center}
\end{table}

\section{Methodology}

Here we followed three innovative approaches in NLP: 1) \texttt{LLaMA 2} fine-tuning with Parameter-Efficient Fine-Tuning (PEFT) techniques such as QLoRA; 2) \texttt{SETFIT} for efficient few-shot fine-tuning of Sentence Transformers; and 3) FAST LSA~\cite{yang2024modeling} V2 on PyABSA framework.

\subsection{LLaMA with QLoRA}
Given the current state-of-the-art interest in Large Language Models (LLMs), we opted to include an LLM-based analysis in our comparative study. 
\texttt{LLaMA 2} is a collection of second-generation open-source LLMs from Meta that comes with a commercial license. \citet{roumeliotis2024llms} presented that \texttt{LLaMA 2} shows a significant leap forward in natural language understanding and generation, by its advanced architecture, large training data and refined training strategies. The architecture of \texttt{LLaMA 2} is based on the transformer model, a neural network architecture that has proven highly effective in a wide range of NLP tasks. \texttt{LLaMA 2} employs a multi-layered transformer architecture with self-attention mechanisms. It is designed to handle a wide range of natural language processing tasks, with models ranging in scale from 7 billion to 70 billion parameters. 

Fine-tuning in machine learning is the process of adjusting the weights and parameters of a pre-trained model on new data to improve its performance on a specific task. 
There are three main fine-tuning methods in the context:
\begin{enumerate}
    \item  \textbf{Instruction Fine-Tunning (IFT):} According to \citet{peng2023instruction}, IFT involves training the model using prompt completion pairs, showing desired responses to queries.
    \item \textbf{Full Fine Tunning:} Full fine-tuning involves updating all of the weights in a pre-trained model during training on a new dataset, allowing the model to adapt to a specific task. 
    \item \textbf{Parameter-Efficient Fine-Tunning (PEFT):} Selectively updates a small set of parameters, making memory requirements more manageable. There are various ways of achieving Parameter efficient fine-tuning. Low-Rank Parameter (LoRA)~\cite{hu2021lora} and Quantized Low-Ranking Adaptation (QLoRA)~\cite{dettmers2023qlora} are the most widely used and effective.
\end{enumerate}

Traditional fine-tuning of pre-trained language models (PLMs) requires updating all of the model's parameters, which is computationally expensive and requires massive amounts of data; thus making it challenging to attempt on consumer hardware due to inadequate VRAMs and computing. However, Parameter-Efficient Fine-Tuning (PEFT) works by only updating a small subset of the model's most influential parameters, making it much more efficient. Four-bit quantization via QLoRA allows such efficient fine-tuning of huge LLM models on consumer hardware while retaining high performance. QLoRA quantizes a pre-trained language model to four bits and freezes the parameters. A small number of trainable \textit{Low-Rank Adapter} layers are then added to the model. In our case, we created a 4-bit quantization with NF4-type configuration using \texttt{BitsAndBytes}\footURL{https://github.com/TimDettmers/bitsandbytes}. 

According to \citet{dettmers2023qlora} under the model fine-tuning process, Supervised fine-tuning (SFT) is a key step in Reinforcement Learning from Human Feedback (RLHF). The SFT models come with tools to train language models using reinforcement learning, starting with supervised fine-tuning, then reward modelling, and finally, Proximal Policy Optimization (PPO). During this process, we provided the SFT trainer with the model, dataset, LoRA configuration, tokenizer, and training parameters.

To test the fine-tuned model, we used the \textit{Transformers} text generation pipeline including the prompt. The \texttt{LLaMA 2} model was fine-tuned using techniques such as QLoRA, PEFT, and SFT to overcome memory and computational limitations. By utilizing Hugging Face libraries such as \texttt {transformers}\footURL{https://huggingface.co/transformers/}, \texttt{accelerate}\footURL{https://huggingface.co/accelerate/}, \texttt{peft}\footURL{https://huggingface.co/peft/}, \texttt{trl}\footURL{https://huggingface.co/trl/}, and \texttt{bitsandbytes}, we were able to successfully fine-tune the 7B parameter \texttt{LLaMA 2} model on a consumer GPU.

\subsection{SetFit}
Few-shot learning has become increasingly essential in addressing label-scarce scenarios, where data annotation is often time-consuming and expensive. These methods aim to adapt pre-trained language models (PLMs) to specific downstream tasks using only a limited number of labelled training examples.
One of the primary obstacles is the reliance on large-scale language models, which typically contain billions of parameters, demanding substantial computational resources and specialized infrastructure. Moreover, these methods frequently require manual crafting of prompts, introducing variability and complexity in the training process, thus restricting accessibility for researchers and practitioners.

In response to this, \citet{tunstall2022efficient} proposed \texttt{SETFIT} (Sentence Transformer Fine-tuning) which presents an innovative framework for efficient and prompt-free few-shot fine-tuning of Sentence Transformers (ST). Diverging from existing methods, \texttt{SETFIT} does not necessitate manually crafted prompts and achieves high accuracy with significantly fewer parameters. %Through a straightforward yet effective approach, \texttt{SETFIT} simplifies the few-shot learning process, making it more accessible and practical across various applications.
The \texttt{SETFIT} approach consists of two main steps. In the first step, the ST is fine-tuned using a contrastive loss function, encouraging the model to learn discriminative representations of similar and dissimilar text pairs. In the second step, a simple classification head is trained on top of the fine-tuned ST to perform downstream tasks such as text classification or similarity ranking. By decoupling the fine-tuning and classification steps, \texttt{SETFIT} achieves high accuracy with orders of magnitude fewer parameters than existing methods, making it computationally efficient and scalable.
In our study, we utilized several available sentence transformers through the \texttt{SETFIT} framework to obtain accuracies for aspect extraction and sentiment polarity identification.

\subsection{PyABSA}
\citet{yang2022pyabsa} addressed the challenge of the lack of a unified framework for ABSA by developing \texttt{PyABSA}, an open-source ABSA framework. \texttt{PyABSA} integrates ATE and text classification functionalities alongside ASC  within a modular architecture. This design facilitates adaptation to various ABSA subtasks and supports multilingual modelling and automated dataset annotation, thereby streamlining ABSA applications.

Moreover, \texttt{PyABSA} offers multi-task-based ATESC models, which are pipeline models capable of simultaneously performing ATE and ASC sub-tasks.% It consists of five main modules: dataset manager, data preprocessor, hyperparameter manager, trainer, and checkpoint manager. This modular structure enables the addition of other sub-tasks or models based on provided templates. 
%Furthermore, \texttt{PyABSA}'s design facilitates easy deployment in a Python environment, making it suitable for ABSA service deployment. 
To tackle the data shortage problem, \texttt{PyABSA} provides automated dataset annotation interfaces and manual dataset annotation tools, encouraging community participation in annotating and contributing custom datasets to the repository.

In our study, we utilized the \texttt{PyABSA} framework on the SemEval 2014 restaurant and laptop benchmark dataset to evaluate accuracy to be consistent with the practices followed in the literature as shown in  \autoref{tab:litAccura}. Specifically, we employed the \texttt{FAST\_LSA\_T\_V2} model with \texttt{PyABSA}, which is included in the \texttt{english} checkpoint, to assess aspect extraction performance. %Using this approach, we conducted an Aspect-based Sentiment Analysis (ABSA) and measured the accuracy of our model. 

\section{Results}

\begin{table*}[!htbp]
%\begin{adjustwidth}{0.1cm}{}
\caption{Accuracies of Models Evaluated by this study on the SemEval 2014~\cite{pontiki-etal-2014-semeval} benchmark. The \textit{Symbol} column is used to refer to the same models in Fig~\ref{fig:graphs} for the sake of brevity.}
\label{tab:Results}
\begin{center}
\small
\begin{tabularx}{0.98\textwidth}{|l|l|c|Y|Y|Y|Y|}
%\begin{tabular}{|l|l|c|c|c|c|c|}
\hline
\multicolumn{2}{|l|}{\multirow{3}{*}{\textbf{Model}}}& \multirow{3}{*}{\parbox[t]{2mm}{\rotatebox[origin=c]{90}{Symbol}}} & \multicolumn{4}{c|}{\textbf{Accuracy (\%)}} \\
\hhline{~~~----}
\multicolumn{2}{|l|}{} & & \multicolumn{2}{c|}{\textbf{Res-14}}& \multicolumn{2}{c|}{\textbf{Lap-14}} \\
\hhline{~~~----}
\multicolumn{2}{|l|}{} & & \makecell{\textbf{Aspect}\\ \textbf{Extraction}} & \makecell{\textbf{Sentiment}\\ \textbf{Polarity}}& \makecell{\textbf{Aspect}\\ \textbf{Extraction}} & \makecell{\textbf{Sentiment}\\ \textbf{Polarity}} \\
\hline
\multicolumn{2}{|l|}{Llama-2-7b~\cite{touvron2023llama} with QLoRA~\cite{dettmers2023qlora}}& - & 35.75 & 65.84 & 71.00 & 65.00\\ 
\hline
\parbox[t]{2mm}{\multirow{17}{*}{\rotatebox[origin=c]{90}{ \texttt{SETFIT}~\cite{tunstall2022efficient}}}} & BGE~\cite{xiao2023c} (Small) & A  & 60.10 & 73.20 & 86.50 & 74.80 \\
\hhline{~------}
& Sentence-T5~\cite{ni2022sentence} (Base) & B & 78.70 & 77.90 & 62.60 & 71.60 \\ 
\hhline{~------}
&   RoBERTa-STSb-v2~\cite{cer2017semeval,reimers2019sentence} (Base) & C & 79.20 & 78.70 & 78.30 & 66.10 \\ 
\hhline{~------}
& Paraphrase-MiniLM-L6-v2~\cite{wang2020minilm,reimers2019sentence} & D & 79.80 & 62.00 & 80.80 & 61.40 \\ 
& ~~+MpNet~\cite{song2020mpnet} & E & 85.40 & 79.50 & 79.40 & 70.00 \\
\hhline{~------}
& CLIP-ViT-B-32-multilingual-v1~\cite{radford2021learning,reimers2019sentence} & F   & 81.90 & 69.30  & 81.70 & 52.60 \\ 
\hhline{~------}
& SPECTER~\cite{specter2020cohan} & G  & 81.90 & 71.60 & 78.60 & 49.60 \\  
\hhline{~------}
& GTR~\cite{ni2022large} (Base) & H & 82.30 & 72.40 & 74.10 & 74.00 \\ 
\hhline{~------}
& TinyBERT~\cite{jiao2020tinybert,reimers2019sentence} & I & 83.10 & 72.40 & 81.20 & 62.90 \\ 
\hhline{~------}
& ALBERT~\cite{lan2019albert,reimers2019sentence} & J  & 76.99 & 71.65 & 77.60 & 65.35 \\ 
& ~~+DistilRoBERTa~\cite{Sanh2019DistilBERTAD} & K  & 84.50 & 75.50 & 82.00 & 66.90 \\ 
\hhline{~------}
& DistilRoBERTa~\cite{Sanh2019DistilBERTAD}& L & 85.00 & 73.20 & 80.80 & 66.10 \\ 
& ~~+All-MiniLM-L6-v2~\cite{wang2020minilm,reimers2019sentence} & M & 85.90 & 71.60 & 77.50 & 65.30 \\  
\hhline{~------}
& MpNet~\cite{song2020mpnet} & N & 87.16 & 77.95 & 87.68 & 70.07 \\ 
\hhline{~------}
& LaBSE~\cite{feng2022language} & O & \textbf{90.30} & 76.40 & 88.40 & 65.40 \\ 
& ~~+MpNet~\cite{song2020mpnet} & P & 88.50 & 74.80 & \textbf{89.50} & \textbf{75.60} \\ 
& ~~+GTR~\cite{ni2022large} (Base) & Q & 88.50 & 74.00 & 87.30 & 73.20 \\ 
& ~~+RoBERTa-STSb-v2~\cite{cer2017semeval,reimers2019sentence} (Base) & R & 88.50 & \textbf{80.30} & \textbf{89.50} & 70.10 \\ 
\hline
\multicolumn{2}{|l|}{FAST LSA~\cite{yang2024modeling} V2 on PyABSA~\cite{yang2023pyabsa}} & - & \multicolumn{2}{c}{\textbf{87.67}} & \multicolumn{2}{|c|}{\textbf{82.60}} \\ 
\hline
%\end{tabular}
\end{tabularx}
\end{center}
%\end{adjustwidth}
\end{table*}

\newcommand{\graphWidth}{0.45}

\begin{figure*}[!htb]
     \centering
     
\begin{subfigure}[!hbt]{\graphWidth\textwidth}
    \centering
    \begin{tikzpicture}
    \begin{axis}[
        xlabel={},
        ylabel={Aspect Extraction Accuracy (\%)},
        xtick=data,
        xticklabels={
            A,
            B,
            C,
            D,
            E,
            F,
            G,
            H,
            I,
            J,
            K,
            L,
            M,
            N,
            O,
            P,
            Q,
            R
        },
        x tick label style={anchor=north,font=\tiny},
        y tick label style={font=\footnotesize},
        ytick={45,50, 55, 60, 65, 70, 75, 80, 85, 90,95,100},
        ymin=45,
        ymax=100,
        width=8cm, % Adjust width here
        height=6cm, % Adjust height here
        grid=both,
        axis lines=left,
        axis line style={-}, % Remove arrow tips from axes
        legend style={at={(0.5,1.85)},anchor=north,font=\footnotesize,yshift=-90pt,draw = none},
        legend columns=3,
        enlarge x limits=0.02,
        ylabel style={font=\scriptsize},
        xlabel style={font=\footnotesize}
    ]
    
    \addplot[mark=*,blue,only marks] coordinates {
        (1, 60.10)
        (2, 78.70)
        (3, 79.20)
        (4, 79.80)
        (5, 85.40)
        (6, 81.90)
        (7, 81.90)
        (8, 82.30)
        (9, 83.10)
        (10, 76.99)
        (11, 84.50)
        (12, 85.00)
        (13, 85.90)
        (14, 87.16)
        (15, 90.30)
        (16, 88.50)
        (17, 88.50)
        (18, 88.50)
    };
    
    \addplot[mark=*,red,only marks] coordinates {
        (1, 86.50)
        (2, 62.60)
        (3, 78.30)
        (4, 80.80)
        (5, 79.40)
        (6, 81.70)
        (7, 78.60)
        (8, 74.10)
        (9, 81.20)
        (10, 77.60)
        (11, 82.00)
        (12, 80.80)
        (13, 77.50)
        (14, 87.68)
        (15, 88.40)
        (16, 89.50)
        (17, 87.30)
        (18, 89.50)
    };

    \legend{Res-14 ,Lap-14}
    
    \end{axis}
    
    % Draw a rectangle around the plot area
    \draw [black] (current axis.south west) rectangle (current axis.north east);
    \end{tikzpicture}
    \caption{Aspect Extraction Accuracy}
    \label{fig:aspect}
\end{subfigure}\hfill
\begin{subfigure}[!hbt]{\graphWidth\textwidth}
    \centering
    \begin{tikzpicture}
    \begin{axis}[
        xlabel={},
        ylabel={Sentiment Polarity Identification Percentage (\%)},
        xtick=data,
        xticklabels={
           A,
            B,
            C,
           D,
           E,
            F,
           G,
           H,
           I,
            J,
            K,
            L,
            M,
            N,
            O,
            P,
            Q,
            R
        },
         x tick label style={anchor=north,font=\tiny},
        y tick label style={font=\footnotesize},
        ytick={45,50, 55, 60, 65, 70, 75, 80, 85, 90,95,100},
        ymin=45,
        ymax=100,
        width=8cm, % Adjust width here
        height=6cm, % Adjust height here
        grid=both,
        axis lines=left,
        axis line style={-}, % Remove arrow tips from axes
        legend style={at={(0.5,1.85)},anchor=north,font=\footnotesize,yshift=-90pt,draw = none},
        legend columns=3,
        enlarge x limits=0.02,
        ylabel style={font=\scriptsize},
        xlabel style={font=\footnotesize}
    ]
    
    \addplot[mark=*,blue,only marks] coordinates {
        (1, 73.20)
        (2, 77.90)
        (3, 78.70)
        (4, 62.00)
        (5, 79.50)
        (6, 69.30)
        (7, 71.60)
        (8, 72.40)
        (9, 72.40)
        (10, 71.65)
        (11, 75.50)
        (12, 73.20)
        (13, 71.60)
        (14, 77.95)
        (15, 76.40)
        (16, 74.80)
        (17, 74.00)
        (18, 80.30)
    };
    
    \addplot[mark=*,red,only marks] coordinates {
        (1, 74.80)
        (2, 71.60)
        (3, 66.10)
        (4, 61.40)
        (5, 70.00)
        (6, 52.60)
        (7, 49.60)
        (8, 74.00)
        (9, 62.90)
        (10, 65.35)
        (11, 66.90)
        (12, 66.10)
        (13, 65.30)
        (14, 70.07)
        (15, 65.40)
        (16, 75.60)
        (17, 73.20)
        (18, 70.10)
    };
    
    \legend{Res-14 ,Lap-14}
    
    \end{axis}
    
    % Draw a rectangle around the plot area
    \draw [black] (current axis.south west) rectangle (current axis.north east);
    \end{tikzpicture}
    \caption{Sentiment Polarity Percentage}
    \label{fig:sentiment_polarity}
    \end{subfigure}
    \caption{Results obtained from SETFIT Models. The models are marked as shown in the \textit{symbols} column in Table~\ref{tab:Results} for brevity.}
    \label{fig:graphs}
\end{figure*}
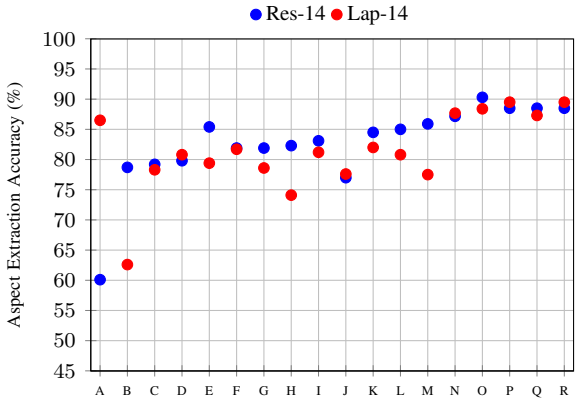
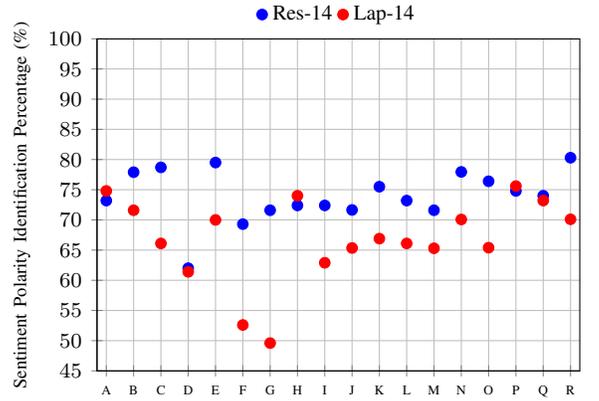

\subsection{LLaMA with QLoRA}
The first section of \autoref{tab:Results} shows the performance of Llama-2-7b~\cite{touvron2023llama} with QLoRA~\cite{dettmers2023qlora}. These performances were obtained using the L4 GPU. It can be noted that even though the sentiment polarity results are comparable between the two datasets, the aspect extraction on \textit{Res-14} is several magnitudes weaker than that of \textit{Lap-14}.

%on the aforementioned dataset by providing accuracy in Aspect extraction and Sentiment Polarity analysis. 

\subsection{SetFit}
% The following section presents a comprehensive analysis of the accuracies achieved by various sentence models utilizing the \texttt{SETFIT} framework. This analysis offers valuable insights into the efficacy and versatility of the \texttt{SETFIT} approach across diverse downstream tasks, including aspect extraction and sentiment polarity identification. 

In the second section of \autoref{tab:Results}, we provide a comprehensive overview of the accuracies attained by various sentence models using the \texttt{SETFIT} framework~\cite{tunstall2022efficient}. Due to the modular nature of the \texttt{SETFIT} framework, we could fine-tune and test combinations of models. If a model is reported in a single row, it means we have used the said sentence transformer model for both aspect extraction and sentiment polarity identification (eg, \texttt{BGE}~\cite{xiao2023c}). In the cell blocks where a model is followed by other models with $+$ are combinations. For example, the first row of \texttt{Paraphrase-MiniLM-L6-v2}~\cite{wang2020minilm,reimers2019sentence} contains results of that model being used both for aspect extraction and sentiment polarity identification. The subsequent line with \texttt{+MpNet}~\cite{song2020mpnet} indicates that \texttt{Paraphrase-MiniLM-L6-v2} was used for the aspect extraction component and \texttt{MpNet} was used for the sentiment polarity identification component. 
%This analysis of sentence model accuracies fine-tuned with \texttt{SETFIT} offers valuable insights into the efficacy and versatility of this approach across diverse downstream tasks. Here with the \texttt{SETFIT} framework, we can either use the same sentence transformer model for both aspect extraction and sentiment polarity identification or two different models to perform the tasks. In \autoref{tab:Results} if it states ``Model 1" under the column Model, it means that the same mentioned model is being used for both the tasks. If it states as ``Model 1 + Model 2" it means that the first mentioned ``Model 1" is being used for aspect extraction and the mentioned ``Model 2" is being used for sentiment polarity identification.

At this point, a question may be raised as to why would the aspect extraction have two different values for accuracy (79.80 vs. 85.40) in the two configurations if in both cases the same model (ie, \texttt{Paraphrase-MiniLM-L6-v2} in this example) was used for that task. The reason is the fact that the fine-tuning is conducted end-to-end in a holistic manner and thus, the choice of the model used for the sentiment polarity identification ends up influencing the ultimate accuracy obtained by the aspect extraction component. It may enhance the result as in the case of \texttt{Paraphrase-MiniLM-L6-v2} and \texttt{MpNet}. It may also hinder as in the case of \texttt{LaBSE}. 

% The presented accuracies offer a detailed glimpse into the efficacy of \texttt{SETFIT}--enhanced models across various tasks within the SemEval 2014 Restaurant and Laptop datasets. 
Overall, it can be noted that \texttt{LaBSE}~\cite{feng2022language} consistently emerges as a standout performer; either by itself or as the aspect extraction component of a pair. It can be argued that this robust performance is owed to its capability to capture nuanced complex information crucial for understanding both aspect-based sentiment analysis and sentiment polarity classification tasks.
%\texttt{LaBSE} consistently emerges as a standout performer, showcasing remarkable accuracy in aspect extraction and sentiment polarity tasks for both datasets. This robust performance underscores \texttt{LaBSE's} capability to capture nuanced complex information crucial for understanding both aspect-based sentiment analysis and sentiment polarity classification tasks.
Specifically on the sentiment polarity classification task, it can be noted that \texttt{Mpnet} and \texttt{RoBERTa-STSb-v2}~\cite{cer2017semeval} elevates performance multiple configurations. Further, the results also reveal domain-specific variations in model performance, as evidenced by the disparity between that \textit{Res-14} and \textit{Lap-14} results of \texttt{SPECTER}~\cite{specter2020cohan}.   
%Furthermore, synergistic combinations with other models such as \texttt{Mpnet and STSB (Roberta-based)} further elevate performance levels, demonstrating the potential for ensemble approaches to enhance model effectiveness. Conversely, models like \texttt{BGE small} and \texttt{paraphrase-MiniLM-L6-v2} exhibit lower accuracies, highlighting the importance of model selection and fine-tuning strategies in achieving optimal performance. The results also reveal domain-specific variations in model performance, as evidenced by \texttt{Allenai-specter's} relatively lower accuracy for sentiment polarity on the Laptop dataset compared to other models. 

%In \autoref{fig:aspect} and \autoref{fig:sentiment_polarity}, the graphs presented provide a comprehensive overview of how various models perform in aspect extraction accuracy and sentiment polarity identification across two distinct domains: SemEval 2014 Restaurant and SemEval 2014 Laptop datasets. By visualizing the performance of these models in different domains, our aim is to provide insights into the robustness of these models for various tasks and datasets.

To give a better overview of how various models perform, we include \autoref{fig:aspect} and \autoref{fig:sentiment_polarity} which visualize the \texttt{SetFit} results discussed in \autoref{tab:Results}. 
In \autoref{fig:aspect}, we present a detailed analysis of aspect extraction accuracy for various models. \texttt{LaBSE} emerging as the top performer across both datasets can easily be noted. It is also evident how \texttt{ALBERT+DistilRoBERTa} and \texttt{LaBSE+RoBERTa-STSb} closely follow with accuracies verging on 90\%. The outlying low accuracies of BGE~\cite{xiao2023c} and Sentence-T5~\cite{ni2022sentence} are also evident.
%In \autoref{fig:aspect}, we present a detailed analysis of aspect extraction accuracy for various models on the SemEval 2014 Restaurant and SemEval 2014 Laptop datasets. Each model's performance is depicted, with \texttt{LaBSE} emerging as the top performer across both datasets. Notably, \texttt{Albert and Distilbert} and \texttt{LaBSE and STSB (Roberta based)} closely follow with accuracies near to 90\%. Conversely, \texttt{BGE small} exhibited consistently lower accuracy compared to other models, underscoring the importance of model selection in aspect extraction tasks.
%
Similarly, in \autoref{fig:sentiment_polarity}, we look into the analysis of sentiment polarity identification percentages for the same models and datasets. Here, \texttt{LaBSE+RoBERTa-STSb} shows the highest accuracy for \textit{Res-14} while \texttt{LaBSE+MpNet} shows the highest accuracy for \textit{Lap-14}. These results reaffirm the effectiveness of \texttt{LaBSE} across both tasks and datasets. Conversely, models such as CLIP-ViT-B-32-multilingual-v1~\cite{radford2021learning} and \texttt{SPECTER} demonstrate relatively lower performances. 
% In \autoref{fig:sentiment_polarity}, we look into the analysis of sentiment polarity identification percentages for the same models and datasets. Here, \texttt{LaBSE and STSB(Roberta based)}showcased the highest accuracy for SemEval 2014 Restaurant, while \texttt{LaBSE} had the highest accuracy for SemEval 2014 Laptop. These results reaffirm the effectiveness of \texttt{LaBSE} across both tasks and datasets. Conversely, models such as \texttt{Allenai-specter} and \texttt{Albert and Distilbert} demonstrated relatively lower performance in sentiment polarity identification.

%Overall, the findings underscore \texttt{SETFIT}'s effectiveness in facilitating efficient few-shot fine-tuning of Sentence Transformers, with \texttt{LaBSE} emerging as a robust choice for aspect extraction and sentiment analysis tasks across diverse datasets.

\subsection{PyABSA}
The third section of \autoref{tab:Results} reports the results obtained from the implementation of the \texttt{FAST LSA}~\cite{yang2024modeling} model on \texttt{PyABSA}~\cite{yang2023pyabsa}. \texttt{PyABSA} is also fine-tuned end-to-end similar to \texttt{SETFIT}~\cite{tunstall2022efficient}. However, unlike \texttt{SETFIT}, \texttt{PyABSA} does not report \textit{Aspect Extraction} and \textit{Sentiment Polarity} accuracies separately. It only gives an overall value. This is the reason for it having only one value per dataset in \autoref{tab:Results}. Alternatively, it is not wrong to take the reported accuracies as the values for the \textit{Sentiment Polarity} task as it is the task that we have at the tail end of the pipe. If regarded in that perspective, it can be claimed that \texttt{FAST LSA} on \texttt{PyABSA} has the best results for \textit{Sentiment Polarity} among all the model combinations and configurations tested by this study. Hence we have opted to highlight those results in bold as we did for the best results in the second (\texttt{SETFIT}) section. 
% The results obtained from the implementation of the \texttt{PyABSA} model are summarized in \autoref{tab:Results}. These accuracies highlight the effectiveness of the \texttt{PyABSA} model in sentiment analysis tasks, particularly in the context of restaurant and laptop reviews.

\section{Conclusion}
This study evaluates three NLP approaches for ABSA: 1) \texttt{LLaMA 2} fine-tuning with Parameter-Efficient Fine-Tuning (PEFT) technique QLoRA; 2) \texttt{SETFIT} for efficient few-shot fine-tuning of Sentence Transformers; and 3) FAST LSA~\cite{yang2024modeling} V2 on PyABSA framework. These approaches aimed to overcome memory and computational limitations while enhancing efficiency and scalability in NLP tasks.
% In conclusion, the paper introduces and evaluates three innovative approaches in natural language processing (NLP): \texttt{LLaMA 2} fine-tuning with techniques like \texttt{QLoRA}, PEFT, and SFT; \texttt{SETFIT} for efficient few-shot fine-tuning of Sentence Transformers; and \texttt{PyABSA} framework for Aspect-Based Sentiment Analysis (ABSA). These approaches aim to overcome memory and compute limitations while enhancing computational efficiency and scalability in NLP tasks.

We observe that \texttt{LLaMA 2}, a collection of second-generation open-source LLMs, after fine-tuning with 4-bit quantization via Parameter-Efficient Fine-Tuning (PEFT) \texttt{QLoRA} only manages middling performance. From the modular options in \texttt{SETFIT}, fine-tuned \texttt{LaBSE} models demonstrate standout performances. Finally, FAST LSA on PyABSA gives out the overall best performance with 87.6\% and 82.6\% accuracy respectively for \textit{Res-14} and \textit{Lap-14} datasets. Nevertheless, none of the tested models are able to surpass the reported accuracy of LSA+DeBERTa-V3-Large~\cite{yang2021improving} which claims 90.33\% and 86.21\% respectively.  
In summary, this study explores the importance of innovative methodologies such as fine-tuning techniques, prompt-free few-shot learning, and modular frameworks in advancing NLP tasks.
%In summary, the paper highlights the importance of innovative methodologies like fine-tuning techniques, prompt-free few-shot learning, and modular frameworks in advancing NLP tasks. These approaches contribute to improved performance, efficiency, and scalability in various NLP applications, offering valuable insights for researchers and practitioners in the field.

\bibliography{bibliography} 
\bibliographystyle{IEEEtranN}

\end{document}